\DocumentMetadata{
  lang = en-US,
  pdfversion = 2.0,
  pdfstandard = ua-2,
  tagging = on,
  tagging-setup = {math/setup=mathml-AF}
}
\documentclass[10pt,twocolumn]{article}

\usepackage[T1]{fontenc}
\usepackage[utf8]{inputenc}
\usepackage{lmodern}
\usepackage{microtype}
\usepackage[margin=0.72in,columnsep=0.24in]{geometry}
\usepackage{amsmath,amssymb}
\usepackage{graphicx}
\usepackage[sort&compress,numbers]{natbib}
\usepackage{xcolor}
\usepackage{tikz}
\usetikzlibrary{arrows.meta,calc,fit,positioning}
\usepackage{hyperref}

\definecolor{acromeldblue}{HTML}{1769AA}
\definecolor{acromeldgreen}{HTML}{167D66}
\definecolor{acromeldorange}{HTML}{B85313}
\hypersetup{
  colorlinks=true,
  linkcolor=acromeldblue,
  citecolor=acromeldgreen,
  urlcolor=acromeldblue,
  bookmarksnumbered=true,
  bookmarksopen=true,
  pdfdisplaydoctitle=true,
  pdftitle={AcroMELD: Recovering Interactive PDF Forms with Structure-Aware Graph Set Transformers},
  pdfauthor={Samuel Abramov},
  pdfsubject={Recovering missing interactive PDF form fields from page images and label-free PDF structure},
  pdfkeywords={PDF forms, form-field detection, document AI, graph set transformer, accessibility},
  pdfcreator={LaTeX}
}

\newcommand{\method}{AcroMELD}
\newcommand{\overlap}{\operatorname{ov}}
\newcommand{\classText}{\textsc{Text}}
\newcommand{\classChoice}{\textsc{Choice}}
\newcommand{\classSignature}{\textsc{Signature}}

\newcommand{\accessibletablerule}{%
  \noalign{%
    \vskip 1pt
    \UseTaggingSocket{tbl/leaders/begin}%
    \hrule height \arrayrulewidth
    \UseTaggingSocket{tbl/leaders/end}%
    \vskip 1pt
  }%
}

\newcommand{\RunCompletedEpochs}{33}
\newcommand{\RunBestRecordIndex}{32}
\newcommand{\RunBestCompletedEpoch}{33}
\newcommand{\TrajectoryNativeFone}{0.8956}
\newcommand{\TrajectoryStrictFone}{0.8564}
\newcommand{\TrajectoryMap}{0.7243}

\newcommand{\ExternalPdfs}{1,996}
\newcommand{\ExternalPages}{6,843}

\newcommand{\InternalMap}{0.6467}
\newcommand{\InternalAPfifty}{0.7792}
\newcommand{\InternalAPseventyfive}{0.7062}
\newcommand{\ExternalStrictMap}{0.2900}
\newcommand{\ExternalStrictAPfifty}{0.4903}
\newcommand{\ExternalStrictAPseventyfive}{0.2772}
\newcommand{\ExternalBaselinePrecision}{0.8338}
\newcommand{\ExternalBaselineRecall}{0.8243}
\newcommand{\ExternalBaselineFone}{0.8290}
\newcommand{\ExternalBaselineTP}{96,070}
\newcommand{\ExternalBaselineFP}{19,145}
\newcommand{\ExternalBaselineFN}{20,478}
\newcommand{\ExternalMargin}{0.0186}
\newcommand{\ExternalSelectedFoneCiLow}{0.8339}
\newcommand{\ExternalSelectedFoneCiHigh}{0.8605}
\newcommand{\ExternalFixedFoneCiLow}{0.7608}
\newcommand{\ExternalFixedFoneCiHigh}{0.7918}
\newcommand{\InternalExternalGap}{0.0867}
\newcommand{\CalibrationGain}{0.0707}
\newcommand{\TextThreshold}{0.769}
\newcommand{\ChoiceThreshold}{0.812}
\newcommand{\SignatureThreshold}{0.990}
\newcommand{\NmsThreshold}{1.000}
\newcommand{\LinkThreshold}{0.950}
\newcommand{\DevSelectedPrecision}{0.9485}
\newcommand{\DevSelectedRecall}{0.9229}
\newcommand{\DevSelectedFone}{0.9355}
\newcommand{\DevSelectedTP}{166,107}
\newcommand{\DevSelectedFP}{9,023}
\newcommand{\DevSelectedFN}{13,870}

\newcommand{\CalibrationSelectedPrecision}{0.9516}
\newcommand{\CalibrationSelectedRecall}{0.9035}
\newcommand{\CalibrationSelectedFone}{0.9269}
\newcommand{\CalibrationSelectedTP}{54,792}
\newcommand{\CalibrationSelectedFP}{2,784}
\newcommand{\CalibrationSelectedFN}{5,854}

\newcommand{\InternalSelectedPrecision}{0.9478}
\newcommand{\InternalSelectedRecall}{0.9214}
\newcommand{\InternalSelectedFone}{0.9344}
\newcommand{\InternalSelectedTP}{443,366}
\newcommand{\InternalSelectedFP}{24,437}
\newcommand{\InternalSelectedFN}{37,824}

\newcommand{\InternalStrictSelectedFone}{0.9042}

\newcommand{\ExternalSelectedPrecision}{0.8592}
\newcommand{\ExternalSelectedRecall}{0.8364}
\newcommand{\ExternalSelectedFone}{0.8477}
\newcommand{\ExternalSelectedTP}{97,486}
\newcommand{\ExternalSelectedFP}{15,974}
\newcommand{\ExternalSelectedFN}{19,062}
\newcommand{\ExternalFixedPrecision}{0.6900}
\newcommand{\ExternalFixedRecall}{0.8890}
\newcommand{\ExternalFixedFone}{0.7770}
\newcommand{\ExternalFixedTP}{103,616}
\newcommand{\ExternalFixedFP}{46,550}
\newcommand{\ExternalFixedFN}{12,932}

\newcommand{\ExternalStrictSelectedFone}{0.7786}

\newcommand{\ExternalStrictFixedFone}{0.7017}

\newcommand{\InternalTextFone}{0.9405}
\newcommand{\ExternalTextPrecision}{0.8455}
\newcommand{\ExternalTextRecall}{0.8247}
\newcommand{\ExternalTextFone}{0.8350}
\newcommand{\ExternalTextAPOverlap}{0.8234}

\newcommand{\ExternalTextStrictAP}{0.5005}
\newcommand{\InternalChoiceFone}{0.9126}
\newcommand{\ExternalChoicePrecision}{0.8870}
\newcommand{\ExternalChoiceRecall}{0.8939}
\newcommand{\ExternalChoiceFone}{0.8904}
\newcommand{\ExternalChoiceAPOverlap}{0.9133}

\newcommand{\ExternalChoiceStrictAP}{0.3782}
\newcommand{\InternalSignatureFone}{0.5498}

\newcommand{\ExternalSignatureRecall}{0.0000}
\newcommand{\ExternalSignatureFone}{0.0000}
\newcommand{\ExternalSignatureAPOverlap}{0.0071}
\newcommand{\ExternalSignatureFixedFone}{0.0677}
\newcommand{\DevSignatureStrictAP}{0.8787}
\newcommand{\InternalSignatureStrictAP}{0.5798}
\newcommand{\ExternalSignatureStrictAP}{0.0023}
\newcommand{\ExternalEmptyPages}{1,439}
\newcommand{\ExternalEmptyPagesWithFP}{242}

\newcommand{\ExternalEmptyPageFPPercent}{16.82}
\newcommand{\ExternalEmptyFPPerPage}{2.6192}
\newcommand{\ExternalSparseFone}{0.7505}
\newcommand{\ExternalMediumFone}{0.8688}
\newcommand{\ExternalDenseFone}{0.9044}

\newcommand{\ExternalPortraitFone}{0.8528}
\newcommand{\ExternalLandscapeFone}{0.8225}
\newcommand{\ExternalSquareFone}{0.9512}
\newcommand{\ExternalSquarePages}{33}
\newcommand{\ExternalVectorPresentFone}{0.8505}
\newcommand{\ExternalVectorAbsentFone}{0.7094}
\newcommand{\ExternalVectorAbsentPages}{388}
\newcommand{\ExternalThroughput}{24.39}
\newcommand{\ExternalPeakMemory}{3.03}

\title{\textbf{AcroMELD: Recovering Interactive PDF Forms with\\
Structure-Aware Graph Set Transformers}}
\author{Samuel Abramov\\
Accessful GmbH\\
\href{mailto:samuel@accessful.de}{\texttt{samuel@accessful.de}}\\
ORCID: \href{https://orcid.org/0009-0007-4511-0739}{0009-0007-4511-0739}}
\date{3 August 2026}

\begin{document}
\maketitle

\begin{abstract}
Interactive PDF form fields are often absent from documents that visually resemble forms,
leaving users unable to enter data without printing or external editing tools.  Detecting the
missing widgets is difficult because a field may be indicated by several overlapping cues,
born-digital PDFs expose useful but incomplete drawing structure, and dense pages can contain
hundreds of fields.  We introduce \method{} (\emph{AcroForm Multi-source Evidence Linking
Decoder}), a 39.4M-parameter detector that combines a high-resolution visual transformer with
label-free PDF primitives.  Its 896-query set comprises 384 visual proposals, 384
structure-seeded proposals, and 128 learned recovery queries.  Four graph-set layers exchange
information over geometry-biased sparse neighborhoods and cross-attend to PDF structure.  A
learned same-field relation links co-referent candidates, while a localization-quality head is
trained on the containment-aware overlap used by the downstream recovery decision.  We define
a hash-bound evaluation protocol with disjoint development, calibration, internal-test, and
quarantined external-holdout roles.  The sealed, single-seed candidate reaches native
containment micro-$F_1$ \InternalSelectedFone{} on the internal test and
\ExternalSelectedFone{} on the one-shot external holdout (95\% PDF-cluster bootstrap interval
\ExternalSelectedFoneCiLow{}--\ExternalSelectedFoneCiHigh{}).  This passes the registered
historical FFGBT-v8 reference by \ExternalMargin{} absolute $F_1$.  Under the stricter external
adapter, however, performance is \ExternalStrictSelectedFone{} IoU-$0.5$ $F_1$ and
\ExternalStrictMap{} COCO mAP, below a locally evaluated CommonForms-L reference; the signature
class receives no prediction at the selected threshold.  Thus the result supports the registered
operational gate while exposing substantial domain and rare-class limitations.
\end{abstract}

\section{Introduction}
\label{sec:introduction}

A PDF can look like a form without containing interactive form controls.  This occurs in
scanned paperwork and in born-digital documents whose lines and boxes were authored only as
page graphics.  The visual appearance is preserved, but the AcroForm widgets needed for direct
data entry are absent.  Recovering candidate widget geometries is therefore a useful first step
toward making such documents fillable.  It is also only one step.  Accessible PDF forms require
keyboard-operable controls, programmatically exposed names, roles, states, and values, a logical
tab and reading order, and appropriate instructions and error feedback
\citep{w3c_pdf3,w3c_pdf12,w3c_pdf22,w3c_pdf23,iso_pdf2}.  We deliberately study the narrower,
measurable problem of localizing and classifying missing field controls.

Recent work casts form-field detection as object detection and demonstrates the importance of
high-resolution page images \citep{barrow2025commonforms}.  A raster, however, discards
information that already exists in a born-digital PDF.  Text runs, line segments, rectangles,
paths, and images carry spatial evidence about candidate fields.  That evidence is neither a
ground-truth annotation nor universally available: a rectangle may be decorative, a text line
may be underlined for emphasis, and a scanned document may expose no useful primitives.
Consequently, PDF structure should complement rather than replace visual evidence.

Combining these sources introduces a set-prediction problem that is easy to overlook.  The same
field can produce a visual proposal, one or more structure-derived proposals, and a learned
recovery proposal.  Standard one-to-one assignment discourages duplicate final predictions
\citep{carion2020detr}, but it does not by itself provide an explicit inference-time statement
that two source-specific candidates refer to the same field.  Dense pages amplify the problem:
our sealed corpus contains a page with 849 valid widgets, well beyond a conventional Top-300
detection cap.  Aggressive non-maximum suppression is also unsafe because neighboring form
controls may legitimately touch or nest.

We propose \method{}, an AcroForm Multi-source Evidence Linking Decoder.  A compact visual
transformer produces 384 proposals, a label-free PDF encoder seeds another 384, and 128 learned
queries recover fields missed by both sources.  The resulting 896-node set is refined by four
graph layers with sparse, geometry-biased self-attention and full cross-attention to the PDF
primitives.  Each node predicts a field class, a box, a containment-aligned localization
quality, and a normalized embedding whose pairwise similarity estimates whether candidates
describe the same field.  Decoding greedily retains representatives and suppresses spatially
eligible candidates inferred to co-refer, without imposing a class-specific duplicate rule.

Our evaluation design treats result provenance as part of the method.  Document groups are
split into five non-overlapping roles; class thresholds and representative-selection parameters
were fit only on calibration data; the candidate was cryptographically frozen before internal
test inference; and a quarantined external corpus was evaluated once after the completed
training lineage validated.  We report both the operational containment criterion and strict
COCO-style localization metrics \citep{lin2014coco}.  Keeping them separate prevents a box that
merely lies inside a target from being mistaken for precise localization.

This work makes four contributions:

\begin{itemize}
  \item A multi-source query formulation that preserves visual, PDF-structural, and unconstrained
  recovery evidence in one exclusive set whose capacity is derived from measured page density.
  \item A geometry-aware graph-set decoder with an explicitly supervised same-field relation,
  allowing candidates to exchange context and be consolidated at inference.
  \item A localization-confidence objective aligned with the containment-aware recovery
  decision, while retaining L1 and generalized-IoU losses and strict COCO evaluation as
  precision safeguards \citep{jiang2018localization,rezatofighi2019giou}.
  \item A leakage-resistant experimental protocol that binds data, code, configuration,
  checkpoints, operating points, and final reports by content hashes.
\end{itemize}

The registered run completed \RunCompletedEpochs{} epochs, froze one candidate and operating
point, evaluated internal test only after that freeze, and then executed the one-shot external
gate.  Its calibrated and fixed-threshold external results differ materially.  With only this
training trajectory, the evidence concerns the complete frozen system rather than the effect of
any one component.

\section{Related Work}
\label{sec:related}

\subsection{Form fields and document layout}
CommonForms formalized field detection as three-class object detection over text inputs, choice
controls, and signature fields, and introduced the high-resolution FFDNet family
\citep{barrow2025commonforms}.  This differs from form understanding datasets such as FUNSD,
where the objective is to connect and semantically interpret text entities
\citep{jaume2019funsd}.  Large layout corpora including PubLayNet and DocLayNet instead annotate
regions such as paragraphs, tables, figures, and headings
\citep{zhong2019publaynet,pfitzmann2022doclaynet}.  These tasks provide valuable document
representations, but their semantic regions do not directly specify interactive widget boxes.
\method{} follows the CommonForms field taxonomy while incorporating low-level PDF primitives
that are normally discarded during rasterization.

\subsection{Multimodal document models}
LayoutLM and later variants jointly encode text, geometry, and image information for document
understanding \citep{xu2019layoutlm,huang2022layoutlmv3}.  DocFormer similarly learns
cross-modal interactions between visual and textual features \citep{appalaraju2021docformer}.
Their usual output is a token-, region-, or document-level semantic prediction.  Our problem
instead requires a high-cardinality, variable-size set of precise boxes and must continue to
operate when text or PDF structure is absent.  We therefore use a small structure encoder as
conditional geometric evidence rather than a document-language backbone.

\subsection{Transformer detection and relational sets}
DETR frames object detection as direct set prediction with bipartite matching
\citep{carion2020detr}; Deformable DETR reduces attention cost using sparse sampling
\citep{zhu2020deformable}.  Our visual branch derives from ECDet-L in EdgeCrafter, a compact-ViT
dense-prediction family \citep{liu2026edgecrafter}.  We retain its COCO-pretrained multiscale
visual proposal generator, then change the downstream problem from a single proposal source to
a heterogeneous query set.

The Set Transformer established attention-based permutation-invariant processing of sets
\citep{lee2018settransformer}.  Relation Networks showed that geometry-aware interactions can
improve recognition and duplicate removal in object detection \citep{hu2017relation}.  Our
graph-set layers draw on both ideas but use the current box hypotheses to define a 32-neighbor
spatial graph.  Unlike generic NMS, a separately supervised link embedding represents
same-field co-reference across query origins and participates directly in representative
selection.

\subsection{Localization confidence}
Classification probability alone need not rank the best localized box.  IoU-Net addressed this
misalignment by predicting localization confidence for ranking and suppression
\citep{jiang2018localization}.  In form recovery, the operational criterion may accept either a
close box or a geometrically contained one.  We train quality against that declared criterion,
but still optimize exact box fit and publish strict IoU metrics.  This dual reporting is
important because containment alone can overvalue an undersized prediction.

\section{Problem Definition and Protocol}
\label{sec:problem}

\subsection{Field-recovery task}

For a PDF page, the model receives a raster image $I$ and an optional sequence of label-free
structure tokens $S$.  Each token represents a text word, line, rectangle, path, image, or the
page itself.  It contains a normalized box, a 32-dimensional primitive descriptor, at most 48
UTF-8 bytes, and an availability flag.  Existing AcroForm widget annotations and field types are
not exposed through $S$.  If extraction fails or the page is scanned, $S$ is a single
unavailable page sentinel.

The target is an unordered set
\begin{equation}
  \begin{aligned}
    Y &= \{(b_j,c_j)\}_{j=1}^{n},\\
    c_j &\in \{\classText,\classChoice,\classSignature\}.
  \end{aligned}
\end{equation}
where $b_j$ is an axis-aligned widget box in page coordinates.  We predict the geometry and
coarse control type needed to instantiate candidate widgets.  We do not predict field names,
descriptions, values, tab order, radio-button groups, required-state semantics, or associations
with instructions.  Accordingly, ``recovery'' in this paper means geometric and type recovery,
not automatic conformance with an accessibility standard.

\subsection{Decision and diagnostic geometry}

The operational adapter inherited from the deployment pipeline uses a symmetric
containment-aware score
\begin{equation}
  \overlap(b,g)=\max\!\left(
    \operatorname{IoU}(b,g),
    \frac{|b\cap g|}{\min(|b|,|g|)}
  \right),
  \label{eq:overlap}
\end{equation}
and accepts a class-consistent match when $\overlap(b,g)\geq 0.3$.  Predictions are processed in
descending score order and greedily matched one-to-one within a page and class.  We call the
resulting precision, recall, and micro-$F_1$ \emph{native} metrics.  This criterion tolerates
different but nested annotations of the same form cue.  It can also accept an undersized box;
therefore every experiment additionally reports strict IoU-$0.5$ operating metrics and COCO
AP averaged over IoU thresholds $0.50{:}0.05{:}0.95$ \citep{lin2014coco}.  The strict adapter
uses a stable, class-global Top-300 cap per page, whereas native evaluation may retain all 896
queries.  Numbers from the two adapters are not interchangeable.

\subsection{Five internal roles and an external gate}
\label{sec:roles}

Partitioning assigns whole document groups to the \emph{train--real}, \emph{train--synthetic},
\emph{development}, \emph{calibration}, and \emph{internal test} roles, so documents that share
provenance never straddle a role boundary.
Weight updates use the two training roles only.  Development native containment micro-$F_1$
selects the exponential-moving-average checkpoint and drives early stopping.  After training,
calibration selects class thresholds and representative-selection parameters.  The resulting
candidate weights and operating point were frozen and bound by digest before internal-test
inference began; that result was used for neither selection nor calibration.

A separate external corpus is quarantined behind a create-last completion record.  Its assets
could not be opened by the evaluation command until the candidate checkpoint, operating point,
run identity, and training-completion hashes all validated.  External document hashes are
excluded from the training index.  The decision reference is FFGBT-v8, the strongest
previously accepted system on this corpus: an unpublished predecessor pipeline of the author
whose sealed external report predates the present project
(Section~\ref{sec:external-references}).  The pre-registered decision was whether the
candidate's native micro-$F_1$ was strictly greater than $0.82903655889853$, reconstructed from
that report's counts $\mathrm{TP}=96{,}070$, $\mathrm{FP}=19{,}145$, and
$\mathrm{FN}=20{,}478$.  The frozen candidate passed that gate in the single permitted external
evaluation.  This decision is specific to the native adapter and does not establish superiority
under the separate strict adapter.

\subsection{Content-addressed provenance}

The run identity binds the sealed SQLite data index, model contract, pretrained checkpoint,
resolved configuration, and an implementation digest covering executable Python, shell, YAML,
and dependency-lock files.  Every checkpoint records those values plus optimizer, scheduler,
mixed-precision, distributed random-number, and early-stopping state.  The candidate was copied
create-once from the selected checkpoint; its digest and those of the operating point, training
summary, metric history, and last checkpoint are bound into a create-only training-completion
record.  The external report was then written atomically with a SHA-256 sidecar and a create-only
completion marker.  Recomputed digests connect the final report to the same candidate and
operating point that produced the post-freeze internal result.  This design does not make an
experiment intrinsically correct, but it prevents a reported checkpoint from silently drifting
away from its registered data or implementation.

\section{AcroMELD}
\label{sec:method}

\begin{figure*}[t]
  \centering
  \begin{tikzpicture}[
    alt={Flow diagram of AcroMELD. A page raster produces 384 visual queries through an ECDet-L
      branch, while PDF primitives produce 384 structure queries through a structure encoder.
      Together with 128 free queries, they enter a four-layer graph set transformer. Per-query
      class, box, quality, and link heads feed representative selection, which outputs text,
      choice, and signature fields. PDF structure also supplies cross-attention memory.},
    node distance=4.0mm and 3.5mm,
    font=\sffamily\footnotesize,
    box/.style args={#1}{draw=#1, very thick, rounded corners=2pt, fill=#1!7,
      align=center, minimum height=9.5mm, inner sep=3pt},
    input/.style={box=gray, text width=21mm},
    branch/.style={box=acromeldblue, text width=23mm},
    query/.style={box=acromeldgreen, text width=23mm},
    core/.style={box=acromeldorange, text width=30mm, minimum height=31mm},
    output/.style={box=acromeldblue, text width=23mm},
    arrow/.style={-{Latex[length=2.1mm]}, thick, draw=gray!75!black},
    cross/.style={-{Latex[length=2.1mm]}, thick, dashed, draw=acromeldorange}
  ]
    \node[input] (raster) {Page raster\\orientation-specific\\letterbox};
    \node[input, below=of raster] (pdf) {PDF primitives\\$\leq512$ tokens\\or sentinel};

    \node[branch, right=of raster] (visual) {ECDet-L visual branch\\COCO initialized\\multiscale features};
    \node[branch, right=of pdf] (structure) {Structure encoder\\2 Transformer layers\\fieldness scores};

    \node[query, right=of visual] (vq) {384 visual queries\\features, boxes, logits};
    \node[query, right=of structure] (sq) {384 structure seeds\\learned fallback slots};
    \node[query, below=of sq] (fq) {128 free queries\\learned grid references};

    \node[core, right=8mm of vq, yshift=-9mm] (graph) {\textbf{Graph Set Transformer}\\896 exclusive queries\\[1.5mm]
      $\times4$ layers:\\32-neighbor geometry attention\\full structure cross-attention\\FFN + iterative box update};

    \node[output, right=of graph, yshift=13mm] (heads) {Per-query heads\\class + no-object\\box + quality\\32-D link embedding};
    \node[output, right=of graph, yshift=-13mm] (select) {Representative selection\\class thresholds\\IoU or learned link\\maximum 896};
    \node[input, right=of select, yshift=13mm] (fields) {Recovered set\\
      {\rmfamily\classText{}}\\{\rmfamily\classChoice{}}\\{\rmfamily\classSignature{}}};

    \draw[arrow] (raster) -- (visual);
    \draw[arrow] (pdf) -- (structure);
    \draw[arrow] (visual) -- (vq);
    \draw[arrow] (structure) -- (sq);
    \draw[arrow] (vq.east) -- (graph.west |- vq.east);
    \draw[arrow] (sq.east) -- (graph.west |- sq.east);
    \draw[arrow] (fq.east) -- (graph.west |- fq.east);
    \coordinate (crossroute) at ($(structure.south)+(0,-23mm)$);
    \coordinate (crossend) at (crossroute -| graph.south);
    \draw[cross] (structure.south) -- (crossroute)
      -- node[below] {cross-attention memory} (crossend) -- (graph.south);
    \draw[arrow] (graph) -- (heads);
    \draw[arrow] (heads) -- (select);
    \draw[arrow] (select) -- (fields);

    \node[draw=gray!60, dashed, rounded corners=3pt, fit=(vq)(sq)(fq),
      inner sep=2.5mm, label={[font=\sffamily\footnotesize,gray!70!black]above:heterogeneous query set}] {};
  \end{tikzpicture}
  \caption{\method{} architecture.  The raster and label-free PDF structure remain independent
  evidence channels until their heterogeneous queries enter one graph-set decoder.  Dashed
  cross-attention makes all structure tokens available as context; solid paths produce the
  exclusive prediction set.}
  \label{fig:architecture}
\end{figure*}
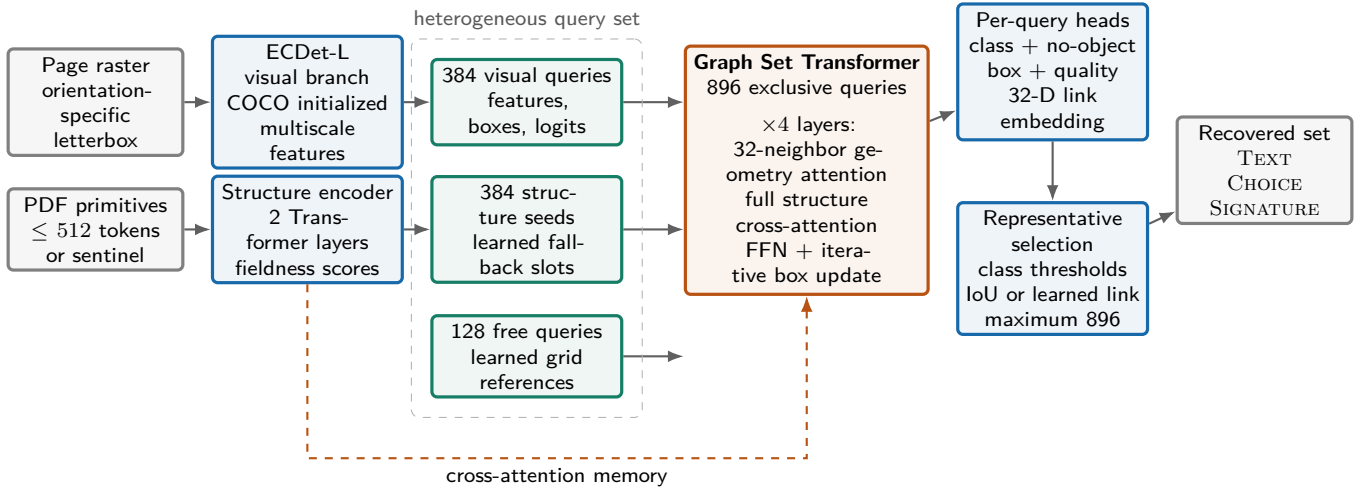

\subsection{Visual and structural evidence}

\subsubsection{Visual proposals}
The raster branch is ECDet-L from EdgeCrafter \citep{liu2026edgecrafter}, adapted from COCO to
three field classes and 384 decoder queries.  It produces proposal features
$V\in\mathbb{R}^{384\times256}$, normalized boxes $B^v$, and class logits $P^v$.  Portrait,
landscape, and approximately square pages are letterboxed without cropping to $2048\times1440$,
$1440\times2048$, and $1664\times1664$ $(H\times W)$ canvases, respectively.  Targets and PDF primitives use
the identical transform.

\subsubsection{Structure encoding}
For token $j$, the structure encoder sums learned kind and availability embeddings, an MLP of
the 32 primitive features, a box embedding, and the mean of its non-padding byte embeddings.
Two pre-normalized Transformer layers contextualize the sequence.  The page token supplies
global context but cannot seed a field.  When raw extraction yields more than 1,535 non-page
primitives, a deterministic first stage preserves 12-by-12 spatial and kind coverage.  A second
deterministic selector retains the page token, prioritizes 16-by-16 kind coverage and form-like
shapes, and caps the encoded sequence at 512 tokens.

A fieldness MLP scores each non-page token from its contextual feature, box, and log aspect
ratio.  The top 384 eligible tokens initialize structure queries.  Missing or ineligible
positions use learned fallback vectors and learned grid reference boxes; therefore the query
shape is constant even when $S$ is unavailable.  Fieldness supervision marks a primitive
positive when its center lies inside any target field.  This signal chooses useful evidence but
does not reveal a target class.

\subsection{Heterogeneous query set}

Let $E_b$ be the box embedding, $E_p$ the projected visual class logits, and $e_o$ a source
embedding.  The initial visual and structural query values are
\begin{align}
  Q^v &= V + E_b(B^v) + E_p(P^v) + e_{\mathrm{visual}},\\
  Q^s &= S_{\mathrm{seed}} + E_b(B^s) + e_{\mathrm{structure}}.
\end{align}
The 128 free values $Q^f$ and their reference boxes are learned on a repeating grid of plausible
field shapes.  Concatenation yields
\begin{equation}
  Q_0=[Q^v;Q^s;Q^f]\in\mathbb{R}^{896\times256}.
\end{equation}
The three sources are not separate prediction heads: after concatenation, Hungarian assignment
operates on one exclusive set.  Capacity 896 follows from the data as the smallest configured
source mixture that exceeds the densest page in the sealed corpus (Section~\ref{sec:data-training}).

\subsection{Geometry-aware graph refinement}

At graph layer $\ell$, each query attends to the $k=32$ query centers nearest to its current box.
For ordered pair $(i,j)$, the attention bias is learned from
\begin{equation}
 \begin{aligned}
 e_{ij}=\bigl[&\Delta x/w_i,\ \Delta y/h_i,\
 \log(w_j/w_i),\ \log(h_j/h_i),\\
 &\operatorname{IoU}_{ij},\ e^{-|\Delta y/h_i|},\ e^{-|\Delta x/w_i|}\bigr].
 \end{aligned}
 \label{eq:edge}
\end{equation}
Eight-head sparse self-attention is followed by eight-head cross-attention from every query to
all encoded structure tokens and a 1,024-dimensional feed-forward block.  Residual connections,
pre-normalization, GELU activations, and dropout 0.1 are used throughout.  The sparse attention
aggregation stores $O(Qk)$ edges; the current exact nearest-neighbor search computes an
$O(Q^2)$ center-distance matrix.  This distinction matters when considering larger future query
sets.

Each of four layers adds a box delta in inverse-sigmoid coordinates, clamps the result to valid
normalized boxes, and emits auxiliary class, box, and quality predictions.  Iterative geometry
therefore changes the graph neighborhood seen by the next layer.

\subsection{Set, quality, and relation objectives}
\label{sec:objectives}

Hungarian assignment uses class, L1-box, and generalized-IoU costs with weights $(2,5,2)$
\citep{carion2020detr,rezatofighi2019giou}.  For matched final queries, the corresponding loss
weights are also $(2,5,2)$, with class weights $(1,1.5,4)$ for \classText{}, \classChoice{}, and
\classSignature{}, and no-object weight 0.08.  Earlier graph layers receive the same set loss
with geometric decay $0.65$.  The native ECDet criterion is retained at weight 0.5 so the visual
proposal interface remains directly supervised.

The quality target for a matched prediction is the detached $\overlap$ value in
Equation~\eqref{eq:overlap}.  Binary cross entropy trains one quality logit $r_i$; background queries are
downweighted as confidence becomes small.  Inference ranks query $i$ by
\begin{equation}
  s_i=\sqrt{\max_c p_{ic}\;\sigma(r_i)},
  \label{eq:score}
\end{equation}
aligning score order with both classification and the declared localization decision.  L1 and
generalized-IoU losses continue to penalize imprecise containment.

The link head maps final queries to unit-normalized 32-dimensional vectors $z_i$.  It predicts
\begin{equation}
  p_{ij}^{\mathrm{link}}=\sigma(z_i^\top z_j/0.1).
\end{equation}
Supervision covers unordered pairs that could interact during decoding: those with
$\operatorname{IoU}\geq0.05$ or $\overlap\geq0.3$.  A pair is positive exactly when both boxes
overlap the same ground-truth field by at least 0.3; class is not part of this label.  Weighted
binary cross entropy handles the positive/negative imbalance.  Relation and fieldness losses
each carry weight 0.2.

\subsection{Containment-aware representative selection}

Candidates below a raw score floor of 0.01 are discarded.  At a calibrated operating point, a
remaining candidate is eligible only if its score meets the threshold for its predicted class.
Eligible candidates then enter a stable descending-score greedy selection.  A selected
representative suppresses candidate $j$ if either their IoU meets or exceeds
$\tau_{\mathrm{nms}}$, or they satisfy the same spatial gate used for relation supervision and
$p_{ij}^{\mathrm{link}}\geq\tau_{\mathrm{link}}$.  Suppression is deliberately class-agnostic:
two sources may disagree on type while still referring to one field.  The selected set is capped
at 896 fields per page.

\section{Data and Training}
\label{sec:data-training}

\subsection{Sealed corpus}

The internal corpus contains 35,388 PDFs, 119,418 pages, and 2,550,287 spatially valid widgets.
Existing native widgets provide supervision, while the model input is rendered without exposing
their metadata through the structure channel.  The index rejects degenerate or off-page boxes
outright, without inventing replacement geometry: 2,295 text, 77 choice, and 798 signature widgets
from source manifests were excluded by this rule.  Splits are made over indivisible provenance
groups and have no content-hash crossing.  The exact post-validation counts are shown in
Table~\ref{tab:data}.

\begin{table*}[t]
  \centering
  \caption{Sealed internal data roles.  ``Structure'' counts pages with extracted PDF primitives;
  pages without structure use the same unavailable sentinel.  Training rows are the only rows
  used for gradient updates.}
  \label{tab:data}
  \small
  \setlength{\tabcolsep}{5.2pt}
  \tagpdfsetup{table/header-rows={1},table/header-columns={1}}
  \begin{tabular}{@{}lrrrrrrr@{}}
    \accessibletablerule
    Role & PDFs & Pages & Structure & \classText{} & \classChoice{} & \classSignature{} & All fields \\
    \accessibletablerule
    Train--real      & 21,272 & 72,548 & 61,332 & 1,356,974 & 319,535 & 6,117 & 1,682,626 \\
    Train--synthetic &  5,000 & 15,785 &      0 &   120,456 &  17,550 & 7,842 &   145,848 \\
    Development      &  2,281 &  7,748 &  7,726 &   145,174 &  34,165 &   638 &   179,977 \\
    Calibration      &    758 &  2,612 &  2,612 &    48,916 &  11,497 &   233 &    60,646 \\
    Internal test    &  6,077 & 20,725 & 19,967 &   388,184 &  91,318 & 1,688 &   481,190 \\
    \accessibletablerule
    Total            & 35,388 &119,418 & 91,637 & 2,059,704 & 474,065 &16,518 & 2,550,287 \\
    \accessibletablerule
  \end{tabular}
\end{table*}

The maximum is 849 fields on one real training page; every split remains below the 896-query
capacity.  Empty pages are retained, including 23,607 real training pages, because false
positive behavior is part of deployment.  Structure is available for 91,637 pages.  The
remainder, including every synthetic page, tests the same visual-only path used for scans.

The 15,785 synthetic pages come from 5,000 generated PDFs.  They are rendered with widgets
disabled, preventing the appearance of a native widget from leaking into its own input image.
Synthetic pages contribute no PDF primitive tokens.  Real signature-bearing pages are sampled
with weight three, while synthetic pages remain at unit weight; this improves rare-class
exposure without allowing the generator distribution to dominate.  The corpus cannot be
redistributed with the code because source-document rights vary; hashes, contracts, and
aggregate statistics are retained for audit instead.

\subsection{Augmentation and missing structure}

Page geometry is preserved.  Images may be scaled by $0.92$--$1.00$ and translated within the
letterbox, followed by brightness/contrast perturbation, blur, JPEG degradation, and sensor
noise.  Flips, mosaics, mixup, arbitrary crops, and rotations are disabled because they can
change form semantics or remove fields.  Raster, boxes, and structure receive one shared
geometric transform.  During training, all structure is dropped on 12\% of pages and 5\% of
non-page tokens are independently dropped; the page sentinel remains.  This creates a declared
visual fallback instead of a hidden zero-filled special case.

\subsection{Optimization}

The registered run uses AdamW with $(\beta_1,\beta_2)=(0.9,0.999)$, weight decay $10^{-4}$,
BF16 autocasting, gradient-norm clipping at 0.1, and exponential-moving-average decay 0.9998.
Learning rates are $2\times10^{-4}$ for the graph/structure modules, $10^{-4}$ for the remaining
visual branch, and $4\times10^{-6}$ for the backbone.  The backbone is frozen for the first two
epochs.  After 1,500 warm-up steps, cosine decay reaches 5\% of each base rate by epoch 30.

Training uses two NVIDIA RTX A6000 GPUs.  An empirical worst-case forward/backward trial on a
$2048\times1440$ page with 849 targets and 512 structure tokens selected four pages per device;
gradient accumulation of four and two distributed ranks produce an effective batch of 32.  The
selected trial allocated 22.10 GiB per GPU; batch eight allocated 44.57 GiB and was rejected by
the reserve policy.  Each completed epoch contains 3,087 optimizer steps.

Main training was registered for at least seven and at most 30 epochs.  Every strict improvement
in development native-containment micro-$F_1$ saved the best EMA checkpoint; patience of five
used a separate minimum increment of 0.0005.  The run reached the 30-epoch main-phase cap without
triggering patience.  The terminal phase then reloaded the best EMA checkpoint, cleared optimizer
state, disabled augmentation, reduced every learning rate by a factor of ten, and completed its
three registered epochs.  Training ended at that terminal cap after \RunCompletedEpochs{} epochs;
the final record, after completed epoch \RunBestCompletedEpoch{}, was also the best EMA checkpoint
and became the frozen candidate.  The registered selection metric was still improving, leaving
convergence unresolved.  All reported results come from this single run, so run-to-run stability
remains unmeasured.

The full model has 39,375,424 trainable parameters.  The visual branch contributes 32,846,939
(83.4\%); the structure encoder and graph-set decoder contribute 6,528,485 (16.6\%).  Parameter
transfer from the published ECDet-L checkpoint covers 99.9969\% of compatible visual parameters;
the three-class score heads are initialized from the mean COCO class weights, while the visual
denoising class embedding is the only compatible visual tensor left random.

\section{Evaluation Design}
\label{sec:evaluation}

\subsection{Operating-point calibration}

Calibration was a discrete, deterministic search performed only after the best EMA checkpoint
and terminal phase were fixed.  The representative-selection grid was the Cartesian product of
\begin{align*}
  \tau_{\mathrm{nms}} &\in \{0.30,0.50,0.70,0.80,0.85,0.90,0.95,1.00\},\\
  \tau_{\mathrm{link}} &\in \{0.50,0.65,0.80,0.90,0.95,1.00\}.
\end{align*}
The $(1,1)$ point is near-unsuppressed rather than suppression-free: the decoder uses inclusive
comparisons, so exact-one IoU or link probabilities may still suppress a candidate.  For each of
the 48 decoder points, class thresholds were selected jointly from 981 descending values between
0.99 and 0.01 at resolution 0.001.  An exact finite fractional optimizer maximized native
containment micro-$F_1$ without materializing the cubic threshold grid.  Integer cross-products
avoided floating-point disagreement; ties preferred higher class thresholds and then the
least-aggressive representative settings.

The selected thresholds were
$(\TextThreshold{},\ChoiceThreshold{},\SignatureThreshold{})$ for \classText{}, \classChoice{},
and \classSignature{}, with
$(\tau_{\mathrm{nms}},\tau_{\mathrm{link}})=(\NmsThreshold{},\LinkThreshold{})$.  Forty of the 48
grid cells (all eight NMS values paired with any of the five registered link thresholds below
1.00) shared the same aggregate native calibration $F_1$ and optimized class thresholds, even
though their retained prediction sets may differ.  The registered tie-break selected the
least-aggressive member of that set.  The eight cells with link threshold 1.00 scored lower.
Because the model weights remained fixed, this grid measures operating-point sensitivity; it cannot
speak to the contribution of relation learning.  The resulting JSON operating point was hash-bound
to the candidate checkpoint before internal-test inference.

\subsection{Internal and external reports}

Internal-test inference began only after the candidate and operating point were sealed, and the
internal test was used for neither checkpoint selection nor calibration.  Its report contains
native micro and per-class counts, precision, recall, and $F_1$, together with strict IoU-$0.5$
counts, COCO mAP, AP$_{50}$, AP$_{75}$, and classwise AP.  Native containment AP was not emitted
for the internal split and is therefore not claimed.  The identical frozen candidate and
operating point were subsequently evaluated once on the external holdout of \ExternalPdfs{}
PDFs and \ExternalPages{} pages.

After the sealed point result, 10,000 percentile-bootstrap resamples clustered by PDF and using
seed 12345 quantified descriptive uncertainty.  The same frozen external predictions also
supported descriptive strata by field-count density, vector-primitive availability, page
orientation, class, and target-empty pages.  Neither the intervals nor the strata affected
selection, calibration, or the registered pass/fail decision.  The sealed detection shards do
not retain query-origin or individual suppression-decision provenance, so those two diagnostics
cannot be reconstructed and are not reported.

\subsection{External corpus, references, and comparability}
\label{sec:external-references}

\subsubsection{Corpus provenance and composition}
The external holdout is the frozen validation benchmark of an unpublished predecessor
form-recovery project by the author.  It predates the present project, was inherited read-only
as a hash-bound snapshot, and is the cleaned revision, at \ExternalPdfs{} documents, of an
earlier 2,290-document collection; every document hash is excluded from the training index
(Section~\ref{sec:roles}).  Ground truth is derived exactly as for the internal corpus: the
documents already carry native AcroForm widgets, whose extracted geometry and coarse type form
the native-adapter targets, while a separately maintained COCO file provides the strict
annotations.  The corpus spans \ExternalPages{} pages with 116,548 native
widgets (77,858 \classText{}, 37,221 \classChoice{}, and 1,469 \classSignature{}) and
116,542 strict annotations.  \ExternalEmptyPages{} pages contain no target field,
\ExternalVectorAbsentPages{} pages expose no vector primitives, and no page exceeds 300
fields.  Page rasters are pre-rendered with interactive widget appearance disabled under the
same rendering contract as training, and structure tokens come from the same label-free
extraction.  The documents are third-party material under mixed licenses, so the corpus itself
cannot be released.  Language, jurisdiction, and document-domain
composition were inherited without further annotation and are characterized only through the
descriptive strata in Section~\ref{sec:results}, a limitation Section~\ref{sec:discussion}
records.

\subsubsection{Native decision reference}
FFGBT-v8 (FFGBT/LightGBM version~8 with SnapNet-E3 refinement) is the strongest previously
accepted system on this benchmark, which is why it was registered as the pass/fail reference
before training began.  Architecturally it is unrelated to \method{}: deterministic candidate
boxes derived from PDF-structure primitives and whitespace analysis are scored by
gradient-boosted trees, accepted boxes pass a refinement network, and output is truncated to
300 detections per page.  Its sealed report on the identical corpus and native adapter records
$\mathrm{TP}=96{,}070$, $\mathrm{FP}=19{,}145$, and $\mathrm{FN}=20{,}478$ (micro-$F_1$
\ExternalBaselineFone{}), with classwise $F_1$ of 0.810 for \classText{}, 0.891 for
\classChoice{}, and 0.121 for \classSignature{}.  The pipeline is unpublished, and its
per-document predictions were not retained; only these aggregate and classwise counts document
it, which precludes any paired candidate--baseline uncertainty estimate
(Section~\ref{sec:results}).

\subsubsection{Strict reference}
The strict adapter's CommonForms-L/FFDNet-L reference is a local evaluation of the released
detector, not the result reported by CommonForms on its own test corpus
\citep{barrow2025commonforms}.  That local run reached micro-$F_1$ 0.8014 at IoU 0.5 and
three-class COCO mAP 0.3410.  These references provide adapter-specific context rather than a
joint ranking: native containment and strict IoU use different annotations, matching rules,
and caps.  We therefore compare \method{} with each reference only under the corresponding
adapter and do not compare the two reference values with one another.

\begin{table}[t]
  \centering
  \caption{Evaluation adapters.  Only the native external row bears the registered decision.}
  \label{tab:adapters}
  \small
  \setlength{\tabcolsep}{3.5pt}
  \tagpdfsetup{table/header-rows={1},table/header-columns={1}}
  \begin{tabular}{@{}llll@{}}
    \accessibletablerule
    Adapter & Match & Per-page cap & Purpose \\
    \accessibletablerule
    Native & $\overlap\geq0.3$ & 896 & selection/gate \\
    Strict operating & IoU$\geq0.5$ & 300 & localization \\
    Strict COCO & IoU sweep & 300 & diagnostic AP \\
    \accessibletablerule
  \end{tabular}
\end{table}

\subsection{Evidence boundary}
\label{sec:ablations}

The study evaluates one registered training trajectory and its frozen system.  It includes
neither retrained component ablations nor post-freeze interventions that remove structure or
link-based selection.  The calibration grid measures decoder sensitivity for fixed weights;
naturally occurring vector-availability and density strata are confounded by document source and
visual characteristics.  Isolating the effects of structure seeds, free queries, relation
supervision, or containment-quality supervision requires a controlled intervention.  The
reported evidence is therefore system-level.

\section{Results}
\label{sec:results}

\subsection{Training trajectory and candidate selection}

\begin{figure*}[t]
  \centering
  \includegraphics[
    width=0.93\textwidth,
    alt={Two-panel line chart of 33 development evaluations. Native F1 rises from 0.760 to
      0.896, strict F1 from 0.724 to 0.856, and COCO mAP from 0.551 to 0.724. Classwise COCO AP
      rises overall for text, choice, and signature fields. The final three augmentation-free
      epochs are shaded.}
  ]{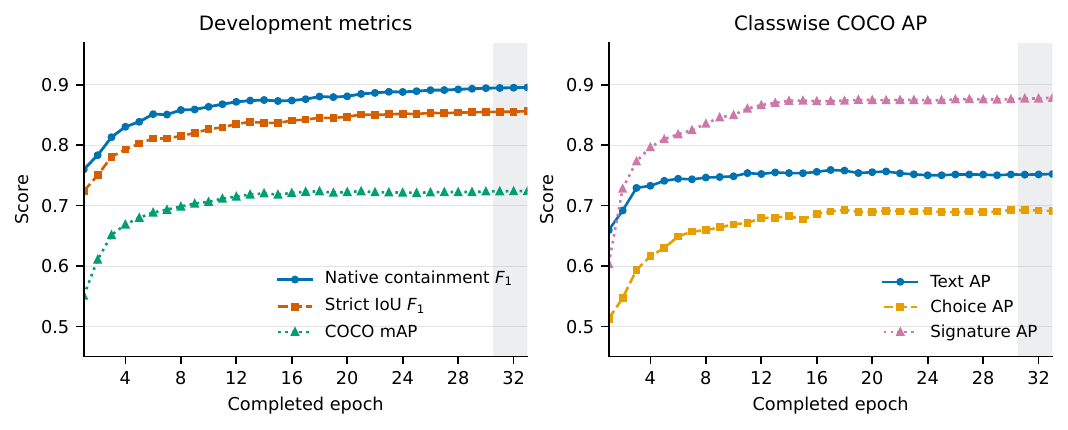}
  \caption{Development trajectory across \RunCompletedEpochs{} completed epochs at the fixed
  training-time decoder point.  Left: native containment $F_1$, strict IoU-$0.5$ $F_1$, and
  strict COCO mAP.  Right: classwise strict COCO AP.  Shading marks the three augmentation-free
  finale epochs.  These repeatedly observed development measurements drove checkpoint selection
  and are not held-out learning curves.}
  \label{fig:dev-curves}
\end{figure*}

The registered run completed \RunCompletedEpochs{} epochs: 30 main-training epochs followed by
three augmentation-free finale epochs.  Figure~\ref{fig:dev-curves} shows the development
measurements used for checkpoint selection.  These measurements retain the training-time decoder
point $(\tau_{\mathrm{nms}},\tau_{\mathrm{link}})=(0.90,0.80)$ and class thresholds
$(0.3,0.3,0.3)$; they are neither calibrated results nor estimates of held-out performance.
Native micro-$F_1$ rose from 0.760 after the first completed epoch to
\TrajectoryNativeFone{} after epoch~\RunBestCompletedEpoch{}, while strict IoU-$0.5$ micro-$F_1$
reached \TrajectoryStrictFone{} and strict COCO mAP reached \TrajectoryMap{}.

The final record, with zero-based index \RunBestRecordIndex{}, was also the selected EMA
checkpoint.  The run reached the pre-registered finale epoch cap while native development $F_1$
had increased for five consecutive epochs.  The development trajectory was therefore still
improving at termination.

\subsection{Calibration and post-freeze evaluation}

Calibration selected class thresholds
$(\TextThreshold{},\ChoiceThreshold{},\SignatureThreshold{})$ for \classText{}, \classChoice{},
and \classSignature{}, respectively, with
$(\tau_{\mathrm{nms}},\tau_{\mathrm{link}})=(\NmsThreshold{},\LinkThreshold{})$.  The selected
native calibration micro-$F_1$ was \CalibrationSelectedFone{}.  The decoder grid was highly
degenerate: all 40 cells with $\tau_{\mathrm{link}}<1$ produced the same aggregate calibration
micro-$F_1$ and selected class thresholds across the complete NMS grid.  The registered tie-break
chose the least-aggressive tied point.  As Section~\ref{sec:evaluation} specifies, this
fixed-weight search selects an operating point but cannot measure the benefit of learned linking.

Table~\ref{tab:native-results} reports the decision-bearing native adapter.  Here and below,
\emph{fixed 0.3} changes only the three class thresholds to $(0.3,0.3,0.3)$; it retains the same
sealed candidate and calibrated NMS/link settings.  Development and calibration rows are included
to disclose their roles; only the internal and external rows are post-freeze estimates.

\begin{table*}[t]
  \centering
  \caption{Native containment results ($\overlap\geq0.3$, maximum 896 detections per page).
  ``Selected'' denotes the calibration-only class thresholds; the fixed-0.3 external control
  changes only those class thresholds.  The FFGBT-v8 row is the registered historical external
  reference used by the pre-registered gate.}
  \label{tab:native-results}
  \small
  \setlength{\tabcolsep}{5.1pt}
  \tagpdfsetup{table/header-rows={1},table/header-columns={1,2}}
  \begin{tabular}{@{}llrrrrrr@{}}
    \accessibletablerule
    Corpus & Point & Precision & Recall & Micro-$F_1$ & TP & FP & FN \\
    \accessibletablerule
    Development & selected & \DevSelectedPrecision{} & \DevSelectedRecall{} &
      \DevSelectedFone{} & \DevSelectedTP{} & \DevSelectedFP{} & \DevSelectedFN{} \\
    Calibration & selected & \CalibrationSelectedPrecision{} & \CalibrationSelectedRecall{} &
      \CalibrationSelectedFone{} & \CalibrationSelectedTP{} & \CalibrationSelectedFP{} &
      \CalibrationSelectedFN{} \\
    Internal test & selected & \InternalSelectedPrecision{} & \InternalSelectedRecall{} &
      \textbf{\InternalSelectedFone{}} & \InternalSelectedTP{} & \InternalSelectedFP{} &
      \InternalSelectedFN{} \\
    External & selected & \ExternalSelectedPrecision{} & \ExternalSelectedRecall{} &
      \textbf{\ExternalSelectedFone{}} & \ExternalSelectedTP{} & \ExternalSelectedFP{} &
      \ExternalSelectedFN{} \\
    External & fixed 0.3 & \ExternalFixedPrecision{} & \ExternalFixedRecall{} &
      \ExternalFixedFone{} & \ExternalFixedTP{} & \ExternalFixedFP{} & \ExternalFixedFN{} \\
    External & FFGBT-v8 & \ExternalBaselinePrecision{} & \ExternalBaselineRecall{} &
      \ExternalBaselineFone{} & \ExternalBaselineTP{} & \ExternalBaselineFP{} &
      \ExternalBaselineFN{} \\
    \accessibletablerule
  \end{tabular}
\end{table*}

At the selected point, the external native micro-$F_1$ of \ExternalSelectedFone{} exceeded the
registered historical FFGBT-v8 value \ExternalBaselineFone{} by \ExternalMargin{}, and therefore
passed the registered one-shot gate.  A 10,000-resample PDF-cluster bootstrap, computed only after the sealed
point result, gives the descriptive interval
$[\ExternalSelectedFoneCiLow{},\ExternalSelectedFoneCiHigh{}]$ for the candidate's external
$F_1$.  This interval describes variation across the evaluated candidate documents; it is not a
confidence interval for the candidate--baseline margin because no corresponding clustered
baseline predictions are available.

The fixed-0.3 control reached only \ExternalFixedFone{}, with descriptive interval
$[\ExternalFixedFoneCiLow{},\ExternalFixedFoneCiHigh{}]$.  Thus class-threshold calibration changed
external $F_1$ by \CalibrationGain{} for the same candidate and representative settings.  Only
aggregate results survive for FFGBT-v8, so the equivalence of its threshold-tuning procedure
cannot be assessed.

Strict results tell a less favorable but complementary story
(Table~\ref{tab:strict-results}).  On the external strict adapter, \method{} reached
micro-$F_1$ \ExternalStrictSelectedFone{} at the selected class thresholds and strict COCO mAP
\ExternalStrictMap{}.  Both its strict operating-point $F_1$ and three-class mAP are below the
local CommonForms-L references of 0.8014 and 0.3410, respectively.  These strict values use
116,542 annotations and a Top-300 cap, placing them outside the native gate comparison.

\begin{table}[t]
  \centering
  \caption{Strict localization diagnostics.  Operating $F_1$ uses IoU$\geq0.5$; mAP uses the
  COCO IoU sweep.  CommonForms-L mAP is the recorded three-class aggregate; comparable
  three-class AP$_{50}$ and AP$_{75}$ aggregates were not recorded.}
  \label{tab:strict-results}
  \small
  \setlength{\tabcolsep}{3.3pt}
  \tagpdfsetup{table/header-rows={1},table/header-columns={1}}
  \begin{tabular}{@{}lrrrr@{}}
    \accessibletablerule
    Corpus / point & $F_1$ & mAP & AP$_{50}$ & AP$_{75}$ \\
    \accessibletablerule
    Internal / selected & \InternalStrictSelectedFone{} & \InternalMap{} &
      \InternalAPfifty{} & \InternalAPseventyfive{} \\
    External / selected & \ExternalStrictSelectedFone{} & \ExternalStrictMap{} &
      \ExternalStrictAPfifty{} & \ExternalStrictAPseventyfive{} \\
    External / fixed 0.3 & \ExternalStrictFixedFone{} & --- & --- & --- \\
    External / CommonForms-L & 0.8014 & 0.3410 & --- & --- \\
    \accessibletablerule
  \end{tabular}
\end{table}

\subsection{Class behavior and descriptive strata}

The aggregate external result conceals a zero-recall outcome on \classSignature{}
(Table~\ref{tab:class-results}).  The selected threshold produced no external signature
predictions, so signature precision is undefined rather than zero; recall and $F_1$ are zero.
Lowering only the class thresholds to 0.3 raised signature $F_1$ to merely
\ExternalSignatureFixedFone{}.  The class-threshold-free strict signature AP also fell from
\DevSignatureStrictAP{} on development to \InternalSignatureStrictAP{} on internal test and
\ExternalSignatureStrictAP{} externally.  The high calibrated threshold directly caused zero
selected predictions.  The weak fixed-threshold $F_1$ and strict AP show that thresholding alone
cannot account for the transfer failure.

\begin{table*}[t]
  \centering
  \caption{Classwise results.  Internal $F_1$ and all external P/R/$F_1$ values use the selected
  native operating point.  External overlap AP and strict COCO AP are class-threshold-free
  ranking diagnostics under their respective adapters.  Signature precision is undefined because no
  signature prediction survived the selected class threshold.}
  \label{tab:class-results}
  \small
  \setlength{\tabcolsep}{5pt}
  \tagpdfsetup{table/header-rows={1},table/header-columns={1}}
  \begin{tabular}{@{}lrrrrrr@{}}
    \accessibletablerule
    Class & Internal $F_1$ & External P & External R & External $F_1$ & Overlap AP & Strict AP \\
    \accessibletablerule
    \classText{} & \InternalTextFone{} & \ExternalTextPrecision{} & \ExternalTextRecall{} &
      \ExternalTextFone{} & \ExternalTextAPOverlap{} & \ExternalTextStrictAP{} \\
    \classChoice{} & \InternalChoiceFone{} & \ExternalChoicePrecision{} &
      \ExternalChoiceRecall{} & \ExternalChoiceFone{} & \ExternalChoiceAPOverlap{} &
      \ExternalChoiceStrictAP{} \\
    \classSignature{} & \InternalSignatureFone{} & --- & \ExternalSignatureRecall{} &
      \ExternalSignatureFone{} & \ExternalSignatureAPOverlap{} &
      \ExternalSignatureStrictAP{} \\
    \accessibletablerule
  \end{tabular}
\end{table*}

The selected native result drops by \InternalExternalGap{} from internal test to external holdout
at the same operating point.  Post-freeze descriptive strata contextualize this gap without
assigning a cause.  External $F_1$ was \ExternalSparseFone{} on pages with 1--10 fields,
\ExternalMediumFone{} with 11--100, and \ExternalDenseFone{} with 101--300; no page in the
external corpus holds more than 300 fields, so a potential dense-page advantage of the 896-query
capacity remains untested.  Pages with vector primitives reached
\ExternalVectorPresentFone{}, compared with \ExternalVectorAbsentFone{} on the
\ExternalVectorAbsentPages{} pages without them.  Because vector availability tracks scan status,
document source, and visual style (Section~\ref{sec:ablations}), the contrast is not a causal
estimate of the structure contribution.

Orientation results were \ExternalPortraitFone{} for portrait and \ExternalLandscapeFone{} for
landscape pages.  The square-page value \ExternalSquareFone{} is based on only
\ExternalSquarePages{} pages.  Of \ExternalEmptyPages{} target-empty pages,
\ExternalEmptyPagesWithFP{} contained at least one false positive
(\ExternalEmptyPageFPPercent{}\%); averaged over all
empty pages, the model emitted
\ExternalEmptyFPPerPage{} false positives per page.  All intervals and strata were computed after
candidate freezing and did not feed back into model, threshold, or gate selection.

\section{Discussion and Limitations}
\label{sec:discussion}

\subsection{Interpreting the registered result}
The sealed candidate passed the pre-registered native gate at its calibration-only operating
point.  That result is specific to the declared adapter and calibration procedure.  With fixed
0.3 class thresholds, the same candidate scored \ExternalFixedFone{}, below FFGBT-v8; only
aggregate baseline results survive, so equivalent threshold tuning cannot be verified.  Under
strict IoU-$0.5$ evaluation, \method{} also remained below the local CommonForms-L reference.
The experiment thus supports the registered operational decision without establishing a general
detector ranking.

Calibration increased external native $F_1$ by \CalibrationGain{} through a precision--recall
trade, including a conservative \SignatureThreshold{} signature threshold.  The 40 tied
NMS/link grid cells identify no preferred suppression setting, and their fixed weights preclude
component attribution.  An explicit no-link intervention or retrained relation ablation is still
needed to test the same-field mechanism.

\subsection{Generalization and signature failure}
The \InternalExternalGap{} internal-to-external $F_1$ gap at one frozen operating point shows that
the internal test is not a reliable proxy for this external domain.  The strongest failure is
\classSignature{}: selected external recall and $F_1$ are zero, and both the fixed-threshold result
and class-threshold-free strict AP remain very low.  Signature-bearing development documents are highly
concentrated in one indivisible provenance group, while synthetic signatures introduce a separate
rendering distribution.  Either could contribute to the gap, but this experiment cannot
distinguish them.  Future work requires provenance-diverse real signatures and group-aware
evaluation rather than further tuning on this holdout.

\subsection{Component and capacity evidence}
Born-digital primitives offer precise geometry without guaranteed semantic relevance.  Dropout
and fallback queries preserve a visual-only path, but the vector-present stratum compares
different document populations; the same pages were never evaluated with and without structure.
The experiment contains no controlled structure-removal or
capacity-matched retraining, so the system-level result cannot isolate the contributions of
structure seeds, free queries, relation supervision, or containment-quality supervision.

Performance increased across the observed positive-page density strata, but the external holdout
never exceeds 300 fields per page.  The 896-query capacity therefore guards against the
truncation measured internally; whether it helps externally over a Top-300 cap cannot be
observed here.

\subsection{Data and metric limitations}
Supervision is derived from existing native widgets.  Such widgets can be inaccurate, duplicated,
or inconsistent with visible cues, and they overrepresent documents that were already
interactive.  The corpus itself cannot be shared (Section~\ref{sec:data-training}), which limits
independent replication of exact training.  Language, jurisdiction,
document-domain, and collection-time distributions have not been comprehensively quantified.
One registered seed provides no estimate of optimization variance.

Containment overlap is useful when two annotations cover the same cue at different extents, yet
its second term can assign a perfect score to a very small box entirely inside a target.  Strict
IoU losses and metrics guard against favorable but imprecise native scores.  Because the two
adapters use different ground truth and caps, the native bootstrap intervals describe only the
candidate under native matching; they do not quantify strict performance or the
candidate--baseline margin.

\subsection{Scope, efficiency, and data responsibility}
Detected boxes and coarse classes do not make a form accessible or semantically correct.  False
positive widgets can obscure content, false negatives preserve an access barrier, and an
incorrect choice/text type changes interaction behavior.  A production system should expose
confidence, preserve the original PDF, and require human review before publication.  Field
labels, descriptions, group semantics, keyboard order, language, error instructions, and
assistive-technology testing remain separate remediation tasks \citep{w3c_pdf23}.  \method{}
does not generate JavaScript or infer document values.

The false positives on target-empty pages illustrate this operational boundary:
\ExternalEmptyPageFPPercent{}\% of such pages received at least one prediction.  A remediation
workflow therefore needs both abstention and human review.

The two-rank external inference pass processed \ExternalPages{} pages at an aggregate
\ExternalThroughput{} pages/s and recorded \ExternalPeakMemory{}~GiB peak allocated GPU memory per
rank on two RTX A6000 GPUs.  This whole-corpus throughput is not batch-one latency.  The model
uses high-resolution canvases, 39.4M parameters, and 896 queries; energy and single-page latency
remain unmeasured.  Pages above 896 fields exceed the training contract and are truncated during
decoding.

PDFs may contain names or other personal information even when publicly obtainable.  The model
does not require target values: structure text is truncated to 48 bytes and used only as an input
feature, while source documents are not distributed with the repository.  This containment does
not replace a full data-protection and licensing review before releasing weights or a service.

\section{Conclusion}
\label{sec:conclusion}

We presented \method{}, a structure-aware graph set transformer for recovering the geometry and
coarse type of missing PDF form widgets.  The method keeps three complementary proposal sources
in one 896-query exclusive set, refines them through sparse geometry-aware interactions and PDF
cross-attention, and learns both containment-aligned localization quality and same-field links.
The experimental protocol binds capacity, adapter semantics, calibration, and artifact lineage
to the reported result.

After \RunCompletedEpochs{} epochs, calibration fixed one operating point before either test
role was opened.  Native micro-$F_1$ reached \InternalSelectedFone{} internally and
\ExternalSelectedFone{} externally, passing the historical FFGBT-v8 gate by \ExternalMargin{}.
However, strict external performance (\ExternalStrictSelectedFone{} IoU-$0.5$ $F_1$ and
\ExternalStrictMap{} COCO mAP) trailed the local CommonForms-L reference, and no signature
prediction survived the selected threshold.  The operational gain therefore coexists with a
clear domain-transfer and rare-class failure.  Further runs and controlled ablations are needed
to determine which architectural components account for the system-level result.

\appendix
\section{Implementation and Run Contract}
\label{app:implementation}

\subsection{Parameter accounting}

\begin{table}[t]
  \centering
  \caption{Trainable parameter allocation.}
  \label{tab:parameters}
  \small
  \tagpdfsetup{table/header-rows={1},table/header-columns={1}}
  \begin{tabular}{@{}lrr@{}}
    \accessibletablerule
    Component & Parameters & Share \\
    \accessibletablerule
    Visual backbone & 22,463,664 & 57.1\% \\
    Other visual modules & 10,383,275 & 26.4\% \\
    Structure encoder & 1,789,184 & 4.5\% \\
    Four graph layers & 4,221,984 & 10.7\% \\
    Query/seed embeddings & 235,649 & 0.6\% \\
    Graph output heads & 281,668 & 0.7\% \\
    \accessibletablerule
    Total & 39,375,424 & 100.0\% \\
    \accessibletablerule
  \end{tabular}
\end{table}

The registered model contract is hidden width 256, eight heads, structure depth two, graph depth
four, feed-forward width 1,024, 384 visual queries, 384 structure queries, 128 free queries, 32
graph neighbors, and link width 32.  There are 23,852 non-trainable buffers.  The query contract
fails closed during training if a page contains more targets than queries.

\subsection{Completed run identity}

The registered run completed 30 main-phase and three augmentation-free finale epochs.  The last
record, zero-indexed epoch 32, was also the selected checkpoint; termination was the registered
\texttt{finale\_epoch\_cap}, not a test-dependent stopping decision.  Training used seed 3407,
two RTX A6000 devices, distributed BF16, per-device batch four, gradient accumulation four, and
effective global batch 32.  No AMP overflow was recorded in any of the 33 complete metric rows.
The principal identities are:
\begin{itemize}
  \item sealed index: \texttt{8d978c02830b\ldots89c20ede};
  \item implementation: \texttt{38b86053e40b\ldots46e6d10};
  \item resolved run configuration: \texttt{fd1519e7a787\allowbreak{}\ldots9c29a2};
  \item model contract: \texttt{8b78a265efd7\ldots56bf20};
  \item pretrained ECDet-L: \texttt{60e48c51b33d\ldots1e85a1}; and
  \item frozen candidate: \texttt{878f94ee9316\ldots{}a8e4dd87}.
\end{itemize}
The implementation digest is reproduced by repository commit
\texttt{227b9a6d651aa882a4776579eb56d15104b8c269}.  The current manuscript checkout contains later
dependency changes and consequently has a different implementation digest.
Appendix~\ref{app:artifacts} records the complete artifact digests and their binding chain.

\subsection{Complete development trajectory}
\label{app:trajectory}

\begin{table}[t]
  \centering
  \caption{All 33 development evaluations.  Rows 1--30 are the main phase and rows 31--33 the
  augmentation-free finale.  The accompanying CSV retains precision, recall, class AP, loss,
  phase, and evaluation time.}
  \label{tab:trajectory}
  \small
  \setlength{\tabcolsep}{5.5pt}
\tagpdfsetup{table/header-rows={1},table/header-columns={1}}
\begin{tabular}{@{}rcccc@{}}
\accessibletablerule
Epoch & Native $F_1$ & Strict $F_1$ & mAP & AP$_{50}$ \\
\accessibletablerule
1 & 0.760 & 0.724 & 0.551 & 0.785 \\
2 & 0.783 & 0.750 & 0.611 & 0.829 \\
3 & 0.813 & 0.781 & 0.651 & 0.853 \\
4 & 0.831 & 0.793 & 0.669 & 0.855 \\
5 & 0.839 & 0.803 & 0.680 & 0.861 \\
6 & 0.851 & 0.811 & 0.689 & 0.865 \\
7 & 0.851 & 0.811 & 0.693 & 0.865 \\
8 & 0.858 & 0.815 & 0.699 & 0.862 \\
9 & 0.859 & 0.820 & 0.703 & 0.864 \\
10 & 0.864 & 0.827 & 0.706 & 0.867 \\
11 & 0.868 & 0.830 & 0.712 & 0.869 \\
12 & 0.872 & 0.835 & 0.715 & 0.869 \\
13 & 0.874 & 0.839 & 0.718 & 0.870 \\
14 & 0.875 & 0.837 & 0.720 & 0.871 \\
15 & 0.873 & 0.837 & 0.718 & 0.869 \\
16 & 0.874 & 0.841 & 0.721 & 0.870 \\
17 & 0.876 & 0.843 & 0.723 & 0.871 \\
18 & 0.881 & 0.845 & 0.724 & 0.870 \\
19 & 0.880 & 0.845 & 0.722 & 0.870 \\
20 & 0.881 & 0.847 & 0.723 & 0.869 \\
21 & 0.885 & 0.851 & 0.724 & 0.869 \\
22 & 0.887 & 0.850 & 0.722 & 0.869 \\
23 & 0.888 & 0.851 & 0.722 & 0.869 \\
24 & 0.888 & 0.852 & 0.722 & 0.869 \\
25 & 0.889 & 0.852 & 0.721 & 0.869 \\
26 & 0.891 & 0.853 & 0.722 & 0.869 \\
27 & 0.891 & 0.853 & 0.723 & 0.869 \\
28 & 0.892 & 0.854 & 0.723 & 0.869 \\
29 & 0.894 & 0.854 & 0.723 & 0.869 \\
30 & 0.894 & 0.855 & 0.724 & 0.869 \\
\accessibletablerule
31 & 0.895 & 0.855 & 0.724 & 0.869 \\
32 & 0.895 & 0.856 & 0.724 & 0.869 \\
33 & 0.896 & 0.856 & 0.724 & 0.870 \\
\accessibletablerule
\end{tabular}

\end{table}

The trajectory uses the registered diagnostic decoder: class thresholds $(0.3,0.3,0.3)$ and
$(\tau_{\mathrm{nms}},\tau_{\mathrm{link}})=(0.90,0.80)$.  The development row in
Table~\ref{tab:native-results} is a separate re-evaluation of the same frozen candidate with the
calibration-selected class thresholds and representative parameters.  Its native $F_1$ is
therefore not directly comparable with the trajectory's final value \TrajectoryNativeFone{}.

\subsection{Decoder controls}

Calibration selected class thresholds
$(\TextThreshold{},\ChoiceThreshold{},\SignatureThreshold{})$ and
$(\tau_{\mathrm{nms}},\tau_{\mathrm{link}})=(\NmsThreshold{},\LinkThreshold{})$.  The search
covered 48 NMS/link cells and 981 class-threshold values per cell.  Forty cells attained the same
aggregate calibration $F_1$ and class thresholds; the registered tie break selected the
least-aggressive corner among those tied cells.  This point is \emph{near-unsuppressed}: inclusive
comparisons still remove identical boxes at IoU 1 or candidates with sufficiently confident
links.  Aggregate ties may also retain different prediction sets.  The search specifies a decoder
operating point; attributing an effect to NMS or link suppression requires an intervention.

\section{Artifact Record, Availability, and Declarations}
\label{app:artifacts}
\label{app:completion}

\newcommand{\artifacthash}[8]{\texttt{#1\allowbreak{}#2\allowbreak{}#3\allowbreak{}#4%
\allowbreak{}#5\allowbreak{}#6\allowbreak{}#7\allowbreak{}#8}}

\subsection{Hash-bound final record}

The create-last completion record binds the frozen candidate to the full metric history, selected
operating point, training summary, resolved configuration, implementation, sealed index, model
contract, and pretrained checkpoint.  The external identity then binds that completed run to the
quarantined corpus and adapter contract; the external report binds itself to that identity, and a
separate completion marker binds the report digest.  The following complete SHA-256 identities
distinguish this result from any later rerun or manuscript revision:
{\scriptsize
\begin{itemize}
  \item \texttt{candidate.pt}:\\
    \artifacthash{878f94ee}{93162807}{109c38f3}{b06bc739}{8b03ba31}{5452ea3c}{7a3b384c}{a8e4dd87}
  \item \texttt{metrics.jsonl}:\\
    \artifacthash{e5b3359d}{444f81e1}{73a8196b}{a1474617}{1dca85ba}{cba7ba71}{728d6cf2}{689627d8}
  \item \texttt{operating-point.json}:\\
    \artifacthash{59961703}{552d3b6f}{5c505008}{981f356b}{3b65f69b}{912ccebf}{0fa49b7d}{6891ee14}
  \item \texttt{training-summary.json}:\\
    \artifacthash{9aa58836}{9c1cb880}{c63f01d1}{6e82f7f6}{9d6d8a50}{0370e543}{17df2bf7}{5c159eab}
  \item \texttt{training-complete.json}:\\
    \artifacthash{df915c6a}{9d85642d}{d207d420}{0f657011}{70fe30b0}{ff908a12}{f192432c}{f4352495}
  \item \texttt{run-identity.json}:\\
    \artifacthash{1bd33553}{ca2fa360}{4f4b323e}{93d4c0db}{3a61a126}{2ade7049}{5602208c}{9da2ca21}
  \item external run identity:\\
    \artifacthash{6f896463}{d5626979}{452e32a3}{bcabcbbb}{32dae657}{5d112626}{dd08caf1}{ddb6994b}
  \item external report:\\
    \artifacthash{903a20a7}{1a7631ae}{b6f2e7e0}{b27ee966}{510b2a7e}{88ee7fdc}{a571040a}{4bde7b84}
  \item external completion marker:\\
    \artifacthash{613335b7}{9d102745}{9603a7da}{82d461b4}{3110320e}{987c9beb}{12643012}{181a3164}
  \item post-freeze analysis:\\
    \artifacthash{4eb25ebb}{9e6d7c00}{8b08de8b}{afbb1096}{971e6213}{404fd105}{198595e3}{b60bfed1}
  \item prediction shard, rank 0:\\
    \artifacthash{1c490926}{971ffe71}{969ba012}{1f3083a7}{defdd23d}{7380416c}{8a7f66c4}{026b16b6}
  \item prediction shard, rank 1:\\
    \artifacthash{fcc9196b}{060867a8}{1fbf3cba}{6923eedc}{259844eb}{a6116082}{d9f7c933}{ab0a6a76}
\end{itemize}}

The repository manifest covers every compact artifact copied into the paper evidence snapshot.
The paper generator pins the manifest contents, recomputes each digest, checks the cross-artifact
bindings above, and fails closed before producing tables or figures.  The report's
\texttt{.sha256} sidecar stores the bare 64-character digest; the two-column
\texttt{SHA256SUMS} file is the canonical input for \texttt{sha256sum -c}.  The post-freeze
analysis is descriptive only: it validates the source PDFs and both prediction shards, then
reconstructs the sealed point counts before calculating PDF-cluster intervals and strata.  It was
not used for selection or calibration.

\subsection{Availability}

The source repository is private at the time of writing and no public archival release or DOI is
claimed.  The trained weights and source PDF corpus are also not publicly distributed.  The
sealed candidate and its verification artifacts (operating point, per-epoch history, external
report, and every seal digest) are held in a private archive and can be made available for
hash audit on request.  The
corpus contains third-party documents with heterogeneous redistribution rights and potentially
sensitive content; releasing either corpus-derived model artifacts or predictions requires a
rights and data-protection assessment.  Thus the hashes, aggregate results, contracts, and
deterministic derivation procedure support audit of the reported record, but an independent exact
rerun is not currently possible from public materials alone.

\subsection{Author contribution and competing interest}

Samuel Abramov designed and implemented the system, curated the experimental protocol, conducted
the experiments, analyzed the results, and wrote the manuscript.  Abramov is affiliated with
Accessful GmbH; because the work concerns document-accessibility technology, this affiliation is
disclosed as a potential competing interest.

{\small
\setlength{\bibsep}{1pt plus 0.3ex}
\bibliographystyle{unsrtnat}
\bibliography{references}
}

\end{document}